\documentclass[10pt, a4paper]{article}

\usepackage[utf8]{inputenc}
\usepackage[T1]{fontenc}
\usepackage{geometry}
\usepackage{graphicx}
\usepackage{amsmath, amssymb}
\usepackage{hyperref}
\usepackage{natbib}
\usepackage{authblk} %

\hypersetup{
    colorlinks=true,
    linkcolor=blue,
    filecolor=magenta,      
    urlcolor=cyan,
    citecolor=blue,
}

\title{kooplearn: A Scikit-Learn Compatible Library of Algorithms for Evolution Operator Learning}

\author[1]{Giacomo Turri}
\author[2]{Grégoire Pacreau}
\author[3]{Giacomo Meanti}
\author[1]{Timothée Devergne}
\author[1]{Daniel Ordonez}
\author[1]{Erfan Mirzaei}
\author[4]{Bruno Belucci}
\author[2]{Karim Lounici}
\author[1,5]{Vladimir Kostic}
\author[1,6]{Massimiliano Pontil}
\author[1]{Pietro Novelli}

\affil[1]{CSML Unit, Italian Institute of Technology, Genoa, Italy}
\affil[2]{CMAP, École Polytechnique, Palaiseau, France}
\affil[3]{Centre Inria de l’Université Grenoble Alpes, Montbonnot, France}
\affil[4]{Paris Dauphine University, Paris, France}
\affil[5]{ Faculty of Science, University of Novi Sad, Serbia}
\affil[6]{Dept. of Computer Science, University College London, U.K.}

\begin{document}

\maketitle

\begin{abstract}
\noindent\texttt{kooplearn} is a machine-learning library that implements linear, kernel, and deep-learning estimators of \textit{dynamical operators} and their spectral decompositions. \texttt{kooplearn} can model both discrete-time evolution operators (Koopman/Transfer) and continuous-time infinitesimal generators. By learning these operators, users can analyze dynamical systems via spectral methods, derive data-driven reduced-order models, and forecast future states and observables. \texttt{kooplearn}'s interface is compliant with the \texttt{scikit-learn} API~\citep{sklearn}, facilitating its integration into existing machine learning and data science workflows. Additionally, \texttt{kooplearn} includes curated benchmark datasets to support experimentation, reproducibility, and the fair comparison of learning algorithms. The software is available at \url{https://github.com/Machine-Learning-Dynamical-Systems/kooplearn}.
\end{abstract}

\section*{Statement of Need}
From fluid flows down to atomistic motions, dynamical systems permeate every scientific discipline. Among the data-driven frameworks for modeling dynamical systems, evolution operator learning~\citep{kostic2022} is both general and principled, and is especially well suited for interpretability~\citep{schutte2001transfer, mezic2005} and dimensionality reduction~\citep{klus2018data}. An evolution operator $\mathsf{E}$ characterizes dynamical systems, either stochastic $x_{t + 1} \sim p(\cdot | x_{t})$, or deterministic $x_{t+1} \sim \delta(\cdot - F(x_{t}))$, as follows: for every function $f$ of the state of the system, $(\mathsf{E} f)(x_{t})$ is the expected value of $f$ one step ahead in the future, given that at time $t$ the system was found in $x_t$
\begin{equation*}
(\mathsf{E} f)(x_t) = \int p(dy | x_{t}) f(y) = \mathbb{E}_{y \sim X_{t + 1} | X_{t}}[f(y) | x_t].
\end{equation*}
Notice that $\mathsf{E}$ is an operator because it maps any function $f$ to another function, $x_{t} \mapsto (\mathsf{E} f)(x_t)$, and is {\em linear} because $\mathsf{E}(f + \alpha g) = \mathsf{E} f + \alpha \mathsf{E} g$. When the dynamics is deterministic, $\mathsf{E}$ is known as the {\em Koopman operator}~\citep{Koopman1931}, while in the stochastic case it is known as the {\em transfer operator}~\citep{Applebaum2009}. Arguably, the most important feature of evolution operators is their spectral decomposition~\citep{mezic2005}, which can be used to express the dynamics as a linear superposition of {\em modes}. These ideas lie at the core of the celebrated Time-lagged Independent Component Analysis~\citep{Molgedey1994}, and  Dynamical Mode Decomposition (DMD)~\citep{Schmid2010, Kutz2016}. 

\begin{figure}[h!]
    \centering
    \includegraphics[width=0.7\textwidth]{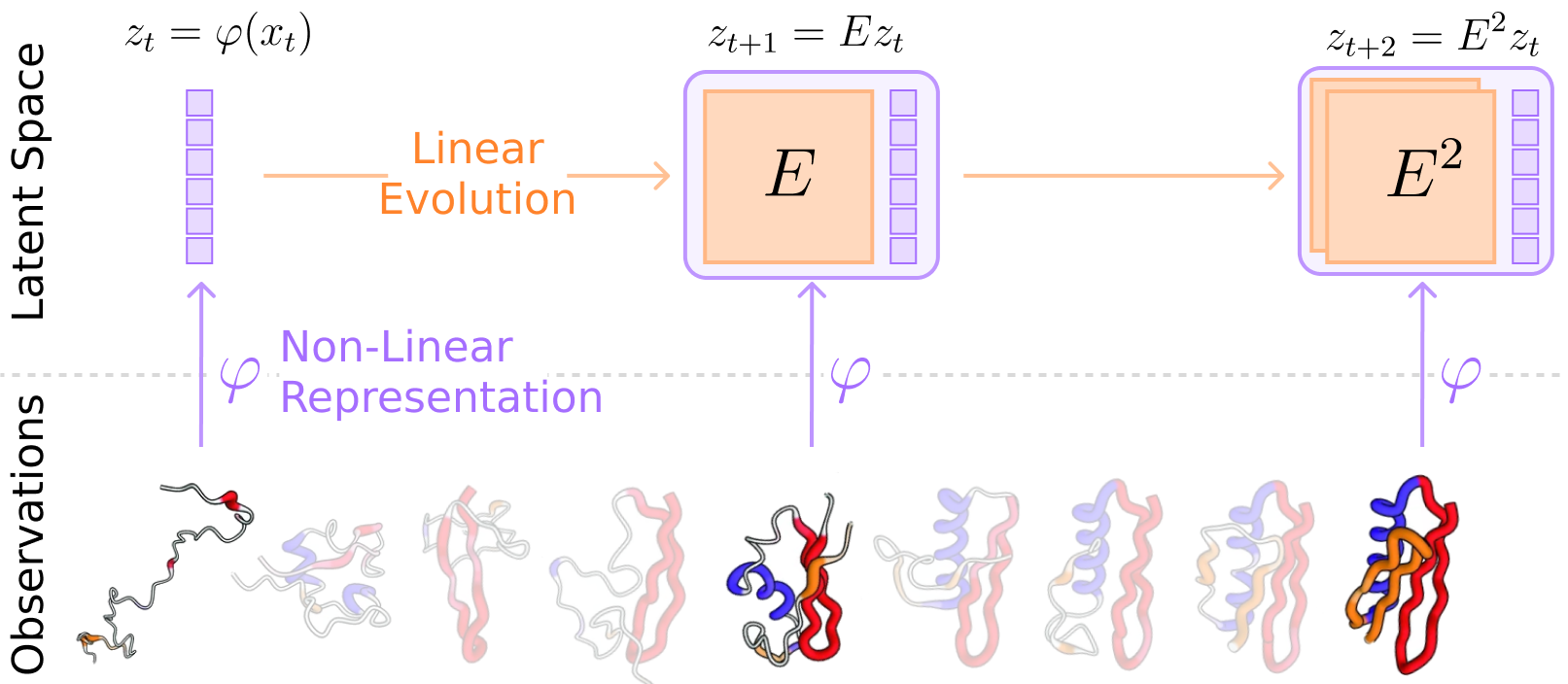}
    \caption{Sketch of the action of an evolution operator on a protein folding trajectory. The dynamics of the protein is linearized by means of a nonlinear representation $\varphi$ and consequently evolved by means of the linear map $E$.}
    \label{fig:evop_scheme}
\end{figure}

Evolution operator learning is best understood from the perspective of {\em latent linear dynamical models}, which is schematically depicted in \autoref{fig:evop_scheme}. In this framework, the dynamical state $x_t$ is first mapped into a latent space defined by a (fixed or learned) representation $\varphi$. Then, a {\em linear evolution} map $E$ is learned to approximate the dynamics of the latents. The pair $(\varphi, E)$ provides an approximation of $\mathsf{E}$ restricted to the $d$-dimensional subspace spanned by the components of $\varphi$, given the data. \texttt{kooplearn} implements state-of-the-art methods to learn $\varphi$, $E$, and the associated spectral decomposition of $\mathsf{E}$. 

\noindent The ecosystem of Python libraries that support operator-based modeling has grown considerably in recent years, with a predominant focus on the DMD family of methods. \texttt{PyDMD} \citep{pydmd} emphasizes classical and kernel DMD variants; \texttt{pykoopman} \citep{pykoopman} implements classical DMD methods with dictionary-based feature maps; \texttt{pykoop} \citep{pykoop} offers a modular framework for lifting-function construction with a focus on system identification and control; \texttt{DLKoopman} \citep{dlkoopman} focuses on autoencoder approaches, while \texttt{KoopmanLab} \citep{koopmanlab} targets Koopman neural operators. \texttt{kooplearn} addresses the general problem of learning evolution operators, and it is the result of a multi-year research effort in innovative operator learning algorithms. While it provides standard prediction and spectral decomposition utilities, it extends the state of the art in evolution operator learning codes by implementing fast kernel estimators \citep{meanti2023,turri2025randomized}, infinitesimal generator models for SDEs \citep{kostic2024generator, devergne2024biased}, and specialized losses for deep representation learning \citep{mardt2018, kostic2024learning, kostic2024neural, turri2025self}. We now provide a concise overview of the functionality of \texttt{kooplearn}.

\subsection*{Learning Linear Evolution Maps $E$}
\begin{figure}[h!]
    \centering
    \includegraphics[width=0.85\textwidth]{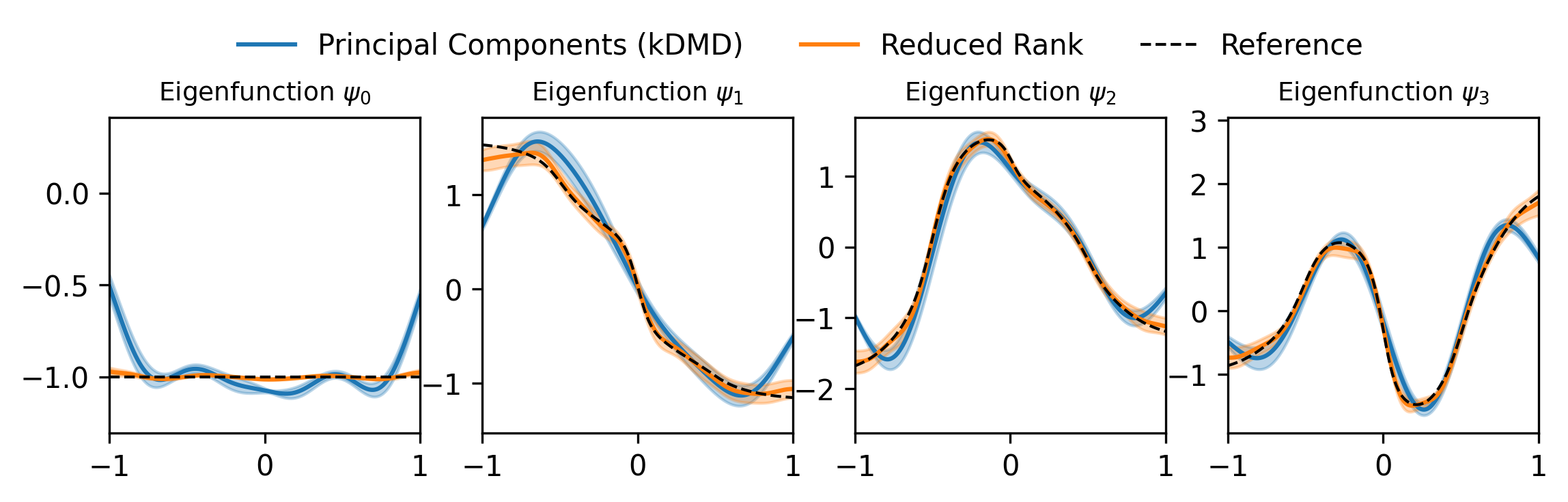}
    \caption{Comparison between kernel DMD (kDMD) and Reduced Rank estimators. The Reduced Rank estimator provides a more accurate approximation of the leading eigenfunctions of the transfer operator for the overdamped Langevin dynamics.}
    \label{fig:eigfns_approximation}
\end{figure}

\begin{figure}[h!]
    \centering
    \includegraphics[width=\textwidth]{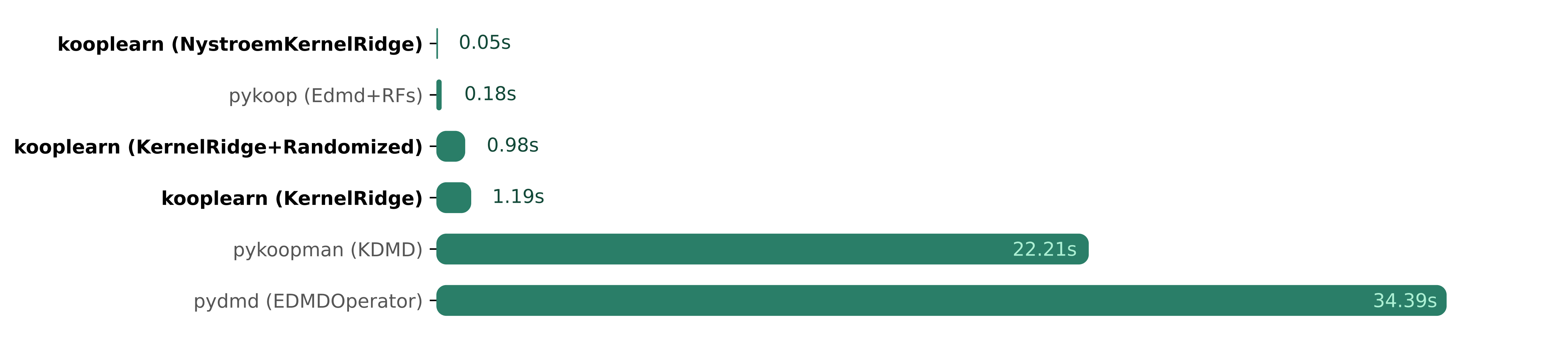}
    \caption{Fit time of a Kernel model (Gaussian kernel) on a dataset of $5000$ observations from the Lorenz 63 dynamical system. The results are the median of three independent runs on a system equipped with an Intel Core i9-9900X CPU (3.50GHz) and 48GB of RAM memory.}
    \label{fig:fast_kernel}
\end{figure}

\texttt{kooplearn} implements state-of-the-art algorithms for learning evolution operators when the representation $\varphi$ is fixed. The library offers estimators in both their linear and kernel formulations (see the \texttt{Ridge} and \texttt{KernelRidge} classes), which bridge the gap between recent theoretical advances~\citep{kostic2022, kostic2023, kostic2024consistent, kostic2024learning} and practical code implementations. A key model in \texttt{kooplearn} is the kernel-based \textit{Reduced Rank Regression}~\citep{kostic2022}. This estimator provably outperforms traditional methods~\citep{williams2015_kdmd} in approximating the operator's spectrum \citep{kostic2023}, as illustrated in \autoref{fig:eigfns_approximation}. To our knowledge, \texttt{kooplearn} provides the only open-source implementation of this algorithm. To handle large datasets, \texttt{kooplearn} also includes randomized~\citep{turri2025randomized} and Nyström-based~\citep{meanti2023} kernel estimators, which significantly speed up the fitting process, making it one of the fastest libraries for kernel-based operator learning, as shown in \autoref{fig:fast_kernel}.

\subsection*{Learning the Representation $\varphi$}
\texttt{kooplearn} also exposes theoretically-grounded loss functions --- implemented in both PyTorch \citep{pytorch} and JAX \citep{jax} --- suited for learning the representation $\varphi$ with neural network models. This allows the incorporation of structural priors, such as graph-based encoders. Within this deep-learning approach, two main families are supported: (i) \textit{encoder-decoder} schemes with the loss proposed in \citet{lusch2018}, and (ii) \textit{encoder-only} schemes, for which \texttt{kooplearn} implements the VAMP loss \citep{mardt2018} and the spectral contrastive loss \citep{turri2025self}.

\subsection*{Learning the Infinitesimal Generator of Diffusion Processes}
In continuous-time dynamics, the system's evolution operator can be expressed as the exponential of the \textit{infinitesimal generator} $\mathsf{L}$, a differential operator defined by the equations of motion~\citep[][Chapter 3]{Applebaum2009}. Formally, for time-homogeneous dynamics, the generator relates to the evolution operator via $\mathsf{E} = e^{\mathsf{L}}$, and consequently $\mathbb{E}[f(X_t)\vert x_0] = (e^{t \mathsf{L}}f)(x_0)$. Since the exponential of an operator preserves its eigenfunctions, one can use knowledge of $\mathsf{L}$ (or its properties) to learn dynamical behavior without requiring lag-time data. In other words, it becomes possible to construct a physics-informed kinetic model $\mathsf{E}$ solely from static (equilibrium) data. To this end, \texttt{kooplearn} provides implementations of recent kernel-based algorithms for diffusion processes with Dirichlet boundary conditions from \cite{kostic2024generator}, as well as neural representations as proposed in \cite{devergne2024biased}. As demonstrated in \cite{devergne2025slow}, these approaches improve sample complexity compared to estimators that rely solely on lag-time trajectory data.

\subsection*{Datasets}
\begin{figure}[h!]
    \centering
    \includegraphics[width=\textwidth]{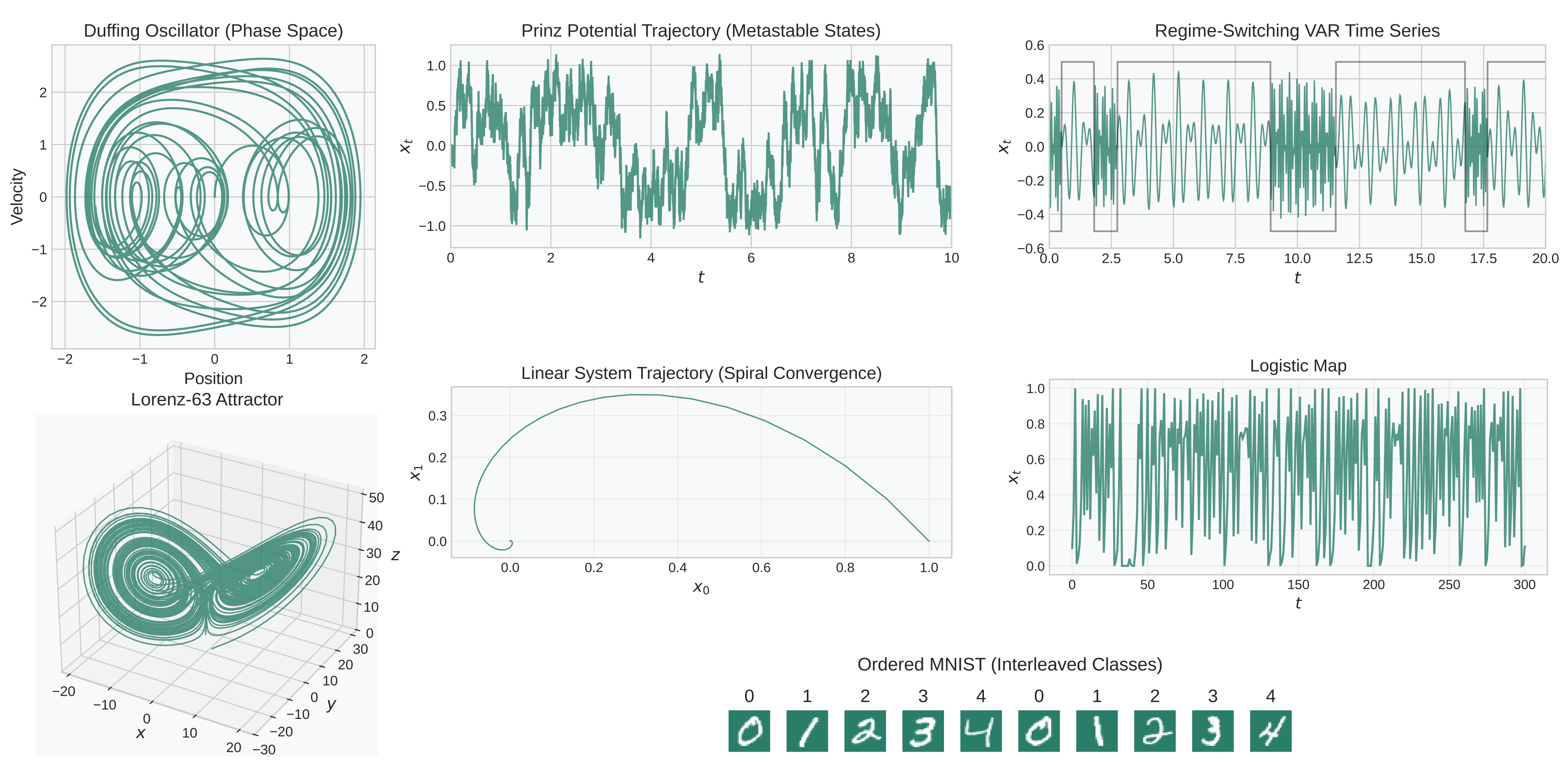}
    \caption{Samples from the datasets included in \texttt{kooplearn}.}
    \label{fig:datasets}
\end{figure}

\noindent To foster reproducibility and rigorous benchmarking, \texttt{kooplearn} includes the \texttt{kooplearn.datasets} module, containing utilities to easily generate trajectories for systems that range from deterministic chaos (e.g., \emph{Lorenz-63}, \emph{Duffing oscillator}, \emph{Logistic Map}) to stochastic and metastable dynamics (e.g., \emph{stochastic linear systems}, \emph{regime-switching models}, \emph{Langevin dynamics}). A distinguishing feature of the library is the inclusion of benchmarks with accessible ground-truth spectral decompositions—such as the \emph{Noisy Logistic Map}~\citep{ostruszka2000dynamical} and \emph{Overdamped Langevin Dynamics} in a quadruple-well potential~\citep{Prinz2011}. These allow users to quantify the accuracy of learned eigenvalues and eigenfunctions directly (as demonstrated in \autoref{fig:eigfns_approximation}). Finally, the suite includes the \emph{Ordered MNIST} from~\citep{kostic2022} to evaluate performance on high-dimensional structured data. Examples of trajectories generated using the \texttt{kooplearn.datasets} module are illustrated in \autoref{fig:datasets}.

\section*{Conclusion}
\texttt{kooplearn} closely follows the \texttt{scikit-learn} API \citep{sklearn} and strives to lower the technical barrier to experimenting with evolution operators. At the same time, it provides optimized implementations of state-of-the-art algorithms for evolution operator learning, making it valuable for research, education, rapid prototyping, and exploratory analysis of dynamical systems. As of today, \texttt{kooplearn} has been employed in a variety of studies \citep{kostic2022, bevanda2023, kostic2023, kostic2024consistent, kostic2024learning, turri2025self, bevanda2025}. It can be installed using the command \texttt{pip install kooplearn}. Its documentation, alongside many worked-out examples, is available on the webpage \url{https://kooplearn.readthedocs.io/}.

\section*{Acknowledgements}

This work was partially funded by the European Union - NextGenerationEU and by the Ministry of University and Research (MUR), National Recovery and Resilience Plan (NRRP), through the PNRR MUR Project PE000013 CUP J53C22003010006 "Future Artificial Intelligence Research (FAIR)" and EU Project ELIAS under grant agreement No. 101120237.

\bibliographystyle{apalike}
\bibliography{paper}

\end{document}